# Deterministic Fitting of Multiple Structures using Iterative MaxFS with Inlier Scale Estimation and Subset Updating

Kwang Hee Lee[2] and Sang Wook Lee[1]
[1]Department of Media Technology, Sogang University, Seoul, Korea
[2]Artificial Intelligence Research Institute, Korea

## Abstract

We present an efficient deterministic hypothesis generation algorithm for robust fitting of multiple structures based on the maximum feasible subsystem (MaxFS) framework. Despite its advantage, a global optimization method such as MaxFS has two main limitations for geometric model fitting. First, its performance is much influenced by the user-specified inlier scale. Second, it is computationally inefficient for large data. The presented MaxFS-based algorithm iteratively estimates model parameters and inlier scale and also overcomes the second limitation by reducing data for the MaxFS problem. Further it generates hypotheses only with top-$n$ ranked subsets based on matching scores and data fitting residuals. This reduction of data for the MaxFS problem makes the algorithm computationally realistic. Our method, called iterative MaxFS with inlier scale estimation and subset updating (IMaxFS-ISE-SU) in this paper, performs hypothesis generation and fitting alternately until all of true structures are found. The IMaxFS-ISE-SU algorithm generates substantially more reliable hypotheses than random sampling-based methods especially as (pseudo-)outlier ratios increase. Experimental results demonstrate that our method can generate more reliable and consistent hypotheses than random sampling-based methods for estimating multiple structures from data with many outliers.

## 1. Introduction

The "hypothesize-and-verify" framework is the core of many robust geometric fitting methods in computer vision. The Random Sample Consensus (RANSAC) algorithm [3] is a widely used robust estimation technique, and most of the state-of-the-art methods are based on random sampling [4, 6, 7, 8, 9, 10, 11, 12, 13, 14, 15, 16, 19]. They involve iterative loop of two steps: random hypotheses generation and verification. A minimal subset of the input data points is randomly sampled and used to hypothesize model parameters. In the verification step, the hypotheses are evaluated against all the data and their support is determined.

There are two main drawbacks to random sampling-based techniques. The first problem is that it is difficult in general to determine the number of iterations to achieve desired confidence without a priori knowledge such as inlier ratio and inlier scale. The true inlier ratio and true inlier scale is usually unknown in most realistic applications. When the number of iterations computed is limited, therefore, the estimated solution may not be reliable. The existence of multiple structures makes the problem more difficult since the inliers belonging to other structures are regarded as outliers (pseudo-outliers).

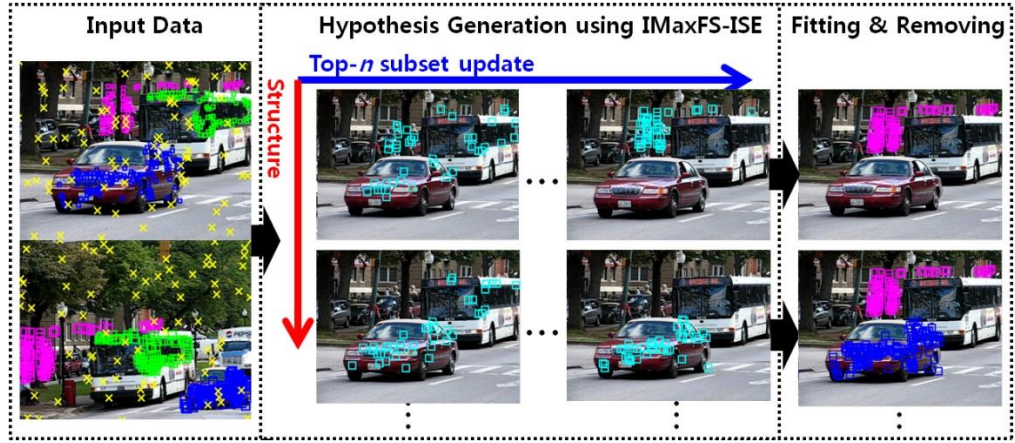

**Fig. 1.** Overview of the presented approach on fundamental matrix estimation.

The second problem is the inconsistency of results, which is related to the first problem. If the number of iterations is insufficient, the random sampling-based techniques provide varying results for the same data and parameter settings. Despite their robustness, the random sampling-based methods provide no guarantee of consistency in their solutions due to the randomized nature [2].

Many advanced random-sampling methods have the same limitations of unreliability and inconsistency. There have been approaches to improving the efficiency of random hypothesis generation for the estimation of single structure [4, 6, 7, 8, 9, 10]. They have been developed to increase the frequency of hitting all-inlier samples.

To deal with multiple-structure data, guided sampling techniques have been developed [11, 12, 13, 14, 15]. They generate a series of tentative hypotheses from minimal subsets of the data in advance and carry out guided sampling based on preference analysis. The performance of these methods can be poor when outlier ratio is considerably high and thus the quality of initial hypothesis is less than desirable. Other multiple-structure model fitting methods also start with random hypothesis generation [16, 17, 18, 5, 19, 14, 20].

Due to the non-deterministic nature of random sampling, the quality of the hypotheses generated by all the methods mentioned above depends on the proportion of pseudo-outliers and gross outliers. Thus, reliable and consistent performance may not be expected when a priori knowledge is not provided. The recent methods such as [21, 22, 23, 24] remove the dependency on inlier scale for the random sampling-based approach. Nevertheless, they cannot overcome the inherent limitation of random sampling.

Deterministic optimization has recently been actively investigated for model fitting problems in computer vision [2, 25, 26, 27, 28]. Despite the guarantee for globally optimal solution, the main limitation of the global optimization algorithms lies in their computational inefficiency. The presence of image features from multiple structures makes their computational cost even higher. Besides, there has been no deterministic method that estimates inlier scale estimation for model fitting problems.

In this paper, we present a deterministic method for robust fitting of multiple structures. The goal of our method is to generate reliable and consistent hypotheses with reasonable efficiency based on the maximum feasible subsystem (MaxFS) framework. There are two limitations in using a MaxFS algorithm for hypothesis generation. First, its performance depends on the user-specified inlier scale. We present an algorithm, called iterative MaxFS with inlier scale (IMaxFS-ISE), that iteratively estimates model parameters and inlier scale. The second limitation is the computational inefficiency mentioned above. We circumvent this limitation by establishing MaxFS problems only with subsets of data but not the whole data. This reduction of data for the MaxFS problem makes our algorithm computationally tractable.

The presented algorithm adopts a sequential "fitting-and-removing" procedure and consists of three major steps: 1) hypothesis generation, 2) labeling and 3) removing inliers. It repeats over these steps until all hypotheses are found. In our method, only one hypothesis is generated for each genuine structure. When a new hypothesis is added, hypothesis selection for each data are performed via the optimization of an energy function and the inliers for each hypothesis generated up to present are removed from whole data. This procedure is repeated until overall energy function is not decreased any more.

The hypothesis generation step itself consists of three steps. The first step is to calculate inlier probability for each set of data. In the second step, input data is sorted according to the inlier probability and the top-n ranked subset is updated. In the last step, the IMaxFS-ISE algorithm estimates the parameters of the hypothesis from the top-n ranked subset. These steps are repeated until the number of inliers is not changed.

Figure 1 provides an overview of our algorithm. In the input images shown on the left side, the yellow crosses indicate the gross outliers, and the other color markers denote three different structures. In the middle of Figure 1, the cyan squares indicate the top-n ranked subset and the images in each row show its update for each structure. The fitting results are shown on the right side of Figure 1.

There have been approaches to using the feature matching scores to increase the chance of finding all-inlier samples [8, 9]. However, they cannot guarantee that correspondences with high matching score are drawn from the same structure. There also has been an approach to the use of fitting residuals for ranking data in selecting a subset for hypothesis generation [12].

Our main contribution is to develop a way of employing a MaxFS algorithm without prior knowledge of inlier ratio, inlier scale and the number of structures. This is unique in that the existing algorithms are predominantly based on random sampling and we believe that the presented method is a viable alternative to the random sampling algorithms. Our algorithm generates substantially more reliable hypotheses than the random sampling methods especially when (pseudo-)outlier ratio is high. This work is an extension of our prior work shown in [38].

The rest of paper is organized as follows: Section 2 introduces our IMaxFS-ISE method. Section 3 describes the algorithm based on fitting-and-removing procedure. Section 4 shows the experimental results with real data, and we conclude in Section 5.

## 2. Iterative Maximum Feasible Subsystem with Inlier Scale Estimation

In this section, we describe main optimization techniques that we employ in our method.

## 2.1. MaxFS Formulation for Geometric Fitting

The aim of a MaxFS framework is to find the largest cardinality set with constraints that are feasible [2, 26]. The objectives of the MaxFS and RANSAC are the same. However, the MaxFS guarantees a global solution unlike the RANSAC. The MaxFS problem admits the mixed integer linear programming (MILP) formulation. The MILP problem is known to be NP-hard. Hence, only relatively small problems can be solved practically. While the exact MILP formulation is useful for small models, it is not effective on large models due to its computational inefficiency [1].

We use the algebraic Direct Linear Transformation (DLT) to estimate hypothesis parameters [29]. The DLT-based geometric fitting problem can be formulated as a MaxFS problem. The set of input data $\mathcal{X}$ is partitioned into the inlier-set $\mathcal{X}^I$ and the outlier-set $\mathcal{X}^O$ with $\mathcal{X}^I \subseteq \mathcal{X}$, $\mathcal{X}^O \subseteq \mathcal{X}$, $\mathcal{X}^I \cup \mathcal{X}^O = \mathcal{X}$ and $\mathcal{X}^I \cap \mathcal{X}^O = \emptyset$.

A maximum inlier scale $s$ provides a bound for the algebraic residual $d_i = |\mathbf{a}_i^T \Theta|$ at point $i$, where $\mathbf{a}_i^T$ is each row vector of $\mathbf{A}$ in the homogeneous equation $\mathbf{A}\Theta = 0$:

$$d_i = |\mathbf{a}_i^T \Theta| \leq s, \ s > 0. \tag{1}$$

A MaxFS formulation of Equation 1 is as follows:

$$\begin{aligned}
\{\hat{\Theta}^{MaxFS}, \hat{\mathbf{y}}\} = \underset{\Theta, \mathbf{y}}{\operatorname{argmin}} & \sum_{i=1}^{k} y_i \\
\text{subject to} \quad & |\mathbf{a}_i^T \Theta| \leq s + M_i y_i, \ \forall i \\
& \mathbf{c}^T \Theta = 1, \\
& \Theta \in \mathfrak{R}^n, \ y_i \in \{0,1\}, \ i = 1,...,k,
\end{aligned} \tag{2}$$

where $M_i$ is a large positive number (Big-M value). The case where $y_i = 0$ indicates that the $i^{th}$ data is an inlier. If $y_i = 1$, the $i^{th}$ data is an outlier and the corresponding constraint is deactivated automatically. We use a linear constraint $\mathbf{c}^T \Theta = 1$, rather than the commonly used $\|\Theta\| = 1$ where $\mathbf{c}$ is a problem dependent vector determined by a user [29]. Our MaxFS algorithm solves Equation 2 for input data $\mathcal{X}$ and the hypothesis $\hat{\Theta}^{MaxFS}$ is generated from the maximum inlier-set.

## 2.2. Iterative MaxFS with Inlier Scale Estimation (IMaxFS-ISE)

The MaxFS problem for geometric fitting can be exactly solved when true inlier scale is known. However, true inlier scale $s$ is unknown in many practical situations, which is commonly set manually by users. In our algorithm, on the other hand, we include a method for estimating the inlier scale from data in a similar iterative manner that has been shown in [24]. A robust scale estimator called IKOSE [24] can estimate the scale of inliers for heavily corrupted multiple-structure data.

Given the true parameters of the $J$th structure ($\theta^J$), the inlier scale $\hat{s}_K^J$ and the inlier number of $J$th structure $v^J$ can be

**Algorithm 1.** Iterative MaxFS with inlier scale estimation

$[\theta^*, s^*, I^{In*}]$ = IMaxFS-ISE(X, M, K)

---

**Input:** input data X, M(for MaxFS), intial inlier scale $s_0$ and K value (for IKOSE)

**Output:** hypothesis parameter $\theta^*$, inlier scale $s^*$ and the number of maximum inliers $I^{In*}$

1: **Initialize** the inlier scale: $s_0$ (small value)
2: **Repeat**
3:   **Estimate** parameter $\theta_t$ and the number of inliers $I^{In}_t$ in X using **MaxFS** with $s_t$. (Sec. 2.1)
4:   **If** $I^{In}_t > I^{In}_{th}$ (We set $I^{In}_{th}$ =10.)
5:     **Estimate** inlier scale $s_{t+1}$ using **IKOSE**(X, K) using Equation (3) (Sec.2.2).
6:   **Else**
7:     $s_{t+1} = s_t + \varepsilon$
8:   **End if**
9:   $\theta^* = \theta_t$,   $I^{In*} = I^{In}_t$   and   $s^* = s_t$
10: **Until** $s_t$ converges

---

estimated by IKOSE. IKOSE for the *J*th structure can be written as follows:

$$\hat{s}_K^J = \frac{|\tilde{r}_K^J|}{\Phi^{-1}(\frac{1}{2}(1+\kappa^J))}, \tag{3}$$

$$\kappa^J := K/v^J, \tag{4}$$

where $|\tilde{r}_K^J|$ is the *k*th sorted absolute residual given the parameters of the *J*th structure ($\theta^J$) and $\Phi^{-1}(\cdot)$ is the argument of the normal cumulative density function and $v^J$ is the number of points satisfying $|r_i^J/\hat{s}_K^J| < 2.5$.

When true inlier scale is known, correct model parameters can be estimated by solving the MaxFS problem. On the other hand, when true model parameters are known, accurate inlier scale can be estimated from IKOSE. To solve this *chicken-and-egg* problem, we develop an iterative scheme which we call IMaxFS-ISE method and it is summarized in Algorithm 1.

We set the initial inlier scale $s_0$ to a small value to guarantee that initial model parameter estimate is not badly biased. In the iteration procedure, the estimated inlier scale $s_t$ increases with the iteration step *t* until it reaches the true inlier scale and the estimated model parameters reach the true model parameters.

## 3. Fitting of Multiple Structures using IMaxFS-ISE and Subset Updating

In this section, we describe our deterministic algorithm for robust fitting of multiple structures. It is summarized in Algorithm 2.

**Algorithm 2.** IMaxFS-ISE-SU Framework

**Input:** input data $\mathcal{X}_N$, M(for MaxFS), initial $K$ value (for IKOSE) $K^{(0)}$ and the number of data points in subset $n$

**Output:** hypotheses parameter set $\Theta = \{\theta^*_l\}^L_{l=1}$ and $S = \{\sigma^*_l\}^L_{l=1}$

1:     $\Theta = \emptyset$, $S = \emptyset$, $\mathcal{X}_{RD} = \mathcal{X}_N$    and    $l = 1$
2:     **Repeat**
3:        $h = 1$ and $K_l^{(h=1)} = K^{(0)}$
4:        **Initialize** top-$n$ ranked subset $\mathcal{X}_n^{(h=1)}$ from $\mathcal{X}_{RD}$
5:        **Repeat**
6:           **Estimate** hypothesis parameter using
            $[\theta_l^{h*}, s_l^{h*}, I^{In}{}_l^{h*}]$=**IMaxFS-ISE**($\mathcal{X}_n^{(h)}$, M, $K_l^{(h)}$) (Sec. 2.2).
7:           **Estimate** inlier scale    $\sigma_l^{h*}$ using IKOSE($\mathcal{X}_N$, $K_l^{(h)}$)
8:           **Calculate** inlier probability $P(x_i)$ (Sec. 3.2)
9:           **Update** top-$n$ ranked subset $\mathcal{X}_n^{(h+1)}$ from $\mathcal{X}_{RD}$
10:          $K_l^{(h+1)} = I^{In}{}_l^{h*}$
11:          $h = h + 1$
12:      **Until** the number of inliers is not changed
13:      $\Theta = \Theta \cup \{\theta_l^{h*}\}$ and $S = S \cup \{\sigma_l^{h*}\}$
14:      **Obtain** labels $f_l$ via α-expansion (Sec. 3.1)
15:      **Generate** the reduced data    $\mathcal{X}_{RD}$ (Sec. 3.2)
16:      $l = l + 1$
17: **Until**    $E(f_{l-1}) < E(f_l)$

## 3.1. Fitting-and-Removing Procedure

Our goal is to estimate the parameters $\Theta = \{\theta_l\}^L_{l=1}$ and the inlier scale $S = \{\sigma_l\}^L_{l=1}$ for multiple structures from input data $\mathcal{X}_N = \{x_i\}^N_{i=1}$. The parameters and inlier scale of the hypothesis for each structure is deterministically estimated using the IMaxFS-ISE. Moreover, only one reliable hypothesis is generated for each structure unlike random sampling-based methods which generate a large number of hypotheses. This facilitates the use of the "fitting-and-removing" procedure.

Our algorithm consists of three major steps: hypothesis generation, labeling and removing inliers. We repeat over these steps until the overall energy function does not decrease. At each iteration stage, a new hypothesis corresponding to a new structure is generated and added to $\Theta$. After a new hypothesis is added, a set of labels $f = \{f_i\}^N_{i=1}$ assigns each data point $x_i$ either to one of the structures or to an outlier by minimizing an objective function using the α-expansion optimization [20]. Our objective function is defined as follows:

$$E(f) = \alpha \sum_{i=1}^{N} D(x_i, f_i) + \beta \sum_{<i,j>\in \mathcal{N}} V(f_i, f_j) + \gamma O(f) \cdot \qquad (5)$$

where $\alpha$, $\beta$ and $\gamma$ are weights of each term. The data cost $D(x_i, f_i)$ in Equation 5 is formulated as

$$D(x_i, f_i) = \begin{cases} r(x_i, \theta_{f_i}) & \text{if } f_i \in \{1,..., L\}, \\ \sigma_{l^*} & \text{if } f_i = 0, \end{cases} \quad (6)$$

$$l^* = \arg\min_l r(x_i, \theta_l), \quad (7)$$

where $r(x_i, \theta_{f_i})$ is the absolute residual of $x_i$ after fitting the structure $\theta_{f_i}$ and $\sigma_{l^*}$ is the inlier scale estimated from the structure $\theta_{l^*}$ which is the penalty for labeling $x_i$ as an outlier. The smoothness cost $V(f_i, f_j)$ in Equation 5 penalizes $f_i \neq f_j$ in some manner. We construct a neighborhood graph from the Delaunay Triangulation on input data $\mathcal{X}_N$ as in [15, 20]. The label cost $O(f)$ is proportional to the number of structures $l$ in $\Theta$ and penalize overly complex models. For the label cost weight, we select a maximum value to include all the true structures. We choose a very small weight (near zero) for the smoothness term since our method estimates a hypothesis with an inlier scale for one structure at a time.

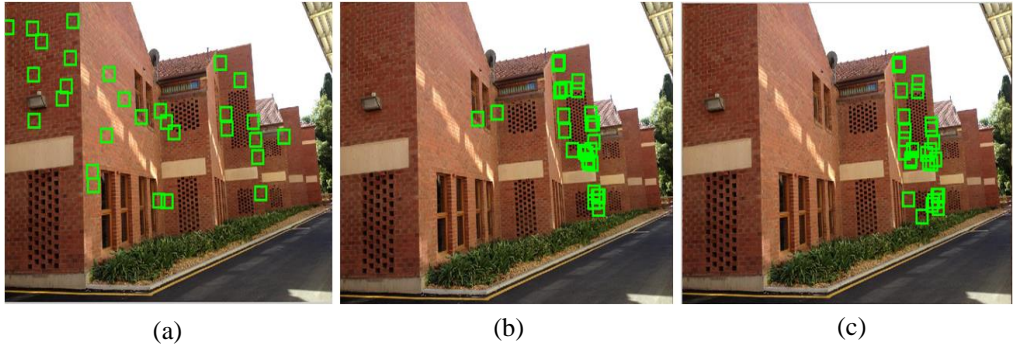

**Fig. 2.** Top-30 subset from the *unihouse* data: (a) initial subset, (b) updated subset after several iterations and (c) updated subset after final iteration

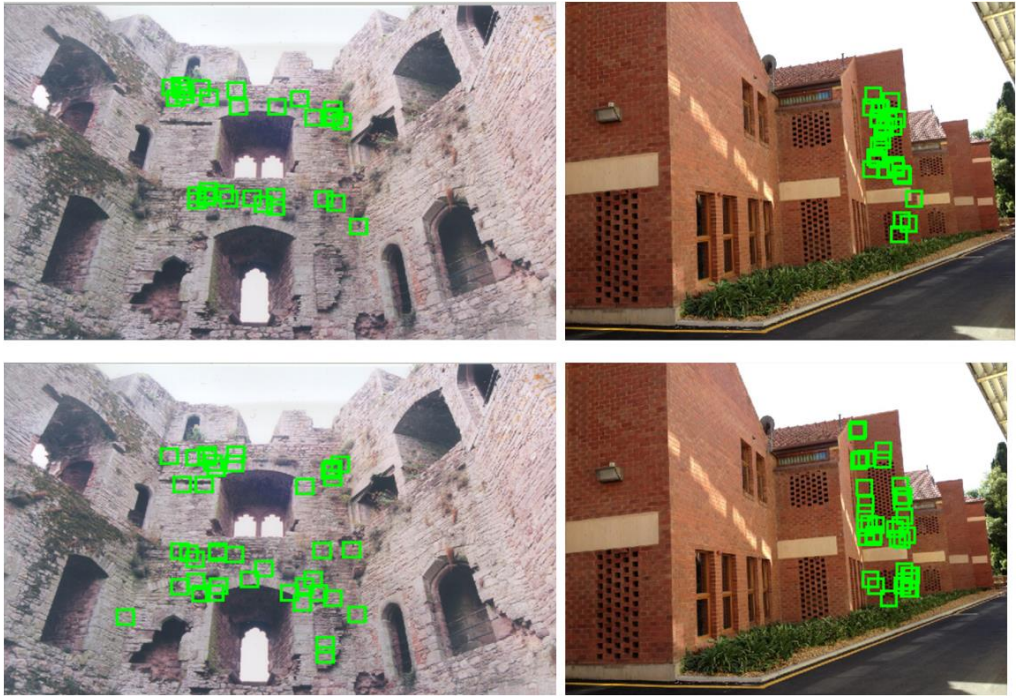

**Fig. 3.** Top-*n* subsets after final iteration: (top) updated using only residuals for ranking and (bottom) updated using both residuals and matching scores.

After labeling is performed for the set of hypotheses $\Theta$ at an iteration stage, a reduced input data $\mathcal{X}_{RD}$ is generated by removing all the estimated inliers from input data $\mathcal{X}_N$. At the next iteration stage, the hypothesis is generated from the reduced input data $\mathcal{X}_{RD}$.

### 3.2. Hypothesis Generation using IMaxFS-ISE from Subset with Top-*n* Ranked Data

We now describe our hypothesis generation method based on the IMaxFS-ISE algorithm. It splits a large problem into smaller ones and thus makes the IMaxFS-ISE algorithm efficient.

Hypothesis generation consists of three steps. The first step is to calculate inlier probability $P(x_i)$ for $x_i$. In the second step, the inlier probability is used to sort the input data $\mathcal{X}_{RD}$ and update the top-*n* ranked subset $\mathcal{X}_n^{(h+1)}$. The last step employs the IMaxFS-ISE algorithm to estimate the parameters of the hypothesis $\theta_l^{h*}$ on the top-*n* ranked subset. These steps are repeated until the number of inliers is not changed.

The initial subset $\mathcal{X}_n^{(0)}$ consists of *n* data with the highest matching scores among the input data $\mathcal{X}_{RD}$. It will contain mostly inliers from the several structures but it is unknown where each inlier belongs. Given $\mathcal{X}_n^{(0)}$, the maximum inliers are estimated with the IMaxFS-ISE algorithm and they are used to generate initial hypothesis.

For a subset, a MaxFS method guarantees that the maximum inliers are found as long as the number of inliers is larger than the minimum number required for estimating model parameters. When the number of structures is large, the inliers from a single structure may be insufficient in the initial subset and the initial hypothesis can significantly deviate from the true structure. When this happens, our algorithm updates the subset using inlier probability until it includes enough inliers from a single structure. Figure 2 shows the initial top-*n* ranked subset, an updated subset after several iterations and the final subset.

Given a hypothesis $\theta_l^h$, we compute the inlier probability of $x_i \in \mathcal{X}_{RD}$ as follows:

$$P(x_i) \propto P(x_i | q) P(x_i | \theta_l^h) = q(x_i) \frac{1}{Z} \exp(\frac{-r(x_i, \theta_l^h)^2}{2\sigma_l^{h^2}}), \quad (8)$$

where $q(x_i)$ is the normalized matching score for the input $x_i$, $r(x_i, \theta_l^h)$ is the absolute residual of $x_i$ computed with the hypothesis $\theta_l^h$ generated for the $l$th structure in the $h$th iteration, $\sigma_l^h$ is the inlier scale corresponding to the hypothesis $\theta_l^h$, and $Z$ is a normalization constant. After the IMaxFS-ISE step is finished in the $h$th iteration, $\sigma_l^h$ is estimated from the whole dataset $\mathcal{X}_N$ using IKOSE. Note that $\sigma_l^h$ is different from the inlier scale $s_l^h$ which is estimated from the subset $\mathcal{X}_n^{(h)}$.

The use of both the inlier scale $\sigma_l^h$ and the matching score $q(x_i)$ in Equation 8 results in more reliable subset than using only residuals for data ranking. When $\theta_l^h$ is badly biased, the inliers of other structures can be included in the top-$n$ ranked subset instead of the outliers with small residuals since $P(x_i)$ is more influenced by $P(x_i | q)$ than $P(x_i | \theta_l^h)$. If $\theta_l^h$ is a good hypothesis, $\theta_l^{h+1}$ can be made better since the $\mathcal{X}_n^{(h+1)}$ includes more inliers. Moreover, the inliers in the subset selected by considering the estimated inlier scale $\sigma_l^h$ tend to be more spread out spatially over the structure. Figure 3 shows an example where inliers are widely distributed in space when the matching scores are used.

One important issue in IMaxFS-ISE is how to choose $K$. To include as many inliers as possible, $K$ should be set to the largest possible value that does not yield breakdown. In our algorithm, $K_l^{(h+1)}$ is set to $I^{In}{}_l^{h*}$ which is the number of maximum inliers estimated from the previous IMaxFS-ISE procedure. On the other hand, we conservatively set the initial value $K^{(0)}$ to a small value, e.g., 10.

## 4. Experimental Results

We have implemented our algorithm in MATLAB using the LP/MILP solver GUROBI [30] which provides functions for the LP/MILP and a desktop with Intel i5-2500 3.30GHz (4 cores) and 3GB RAM is used for experiments. We tested five methods including ours on several real datasets. For performance evaluation and comparison, we measured the actual elapsed computation time. Images and keypoint correspondences were acquired from the Oxford VGG dataset (*mc3* and *raglan*) [31], the Hopkins 155 dataset (*carsbus*) [36] and the Adelaide RMF dataset (*biscuitbookbox, BCD, dinobooks, library, unihouse, bonhall, cubetoy, 4B* and *5B*) [34, 35]. We used manually labeled keypoint correspondences which were obtained by SIFT matching. If keypoint matching scores are not available, we assigned a proper matching score to each correspondence.

### 4.1. IMaxFS-ISE : Line Fitting Results

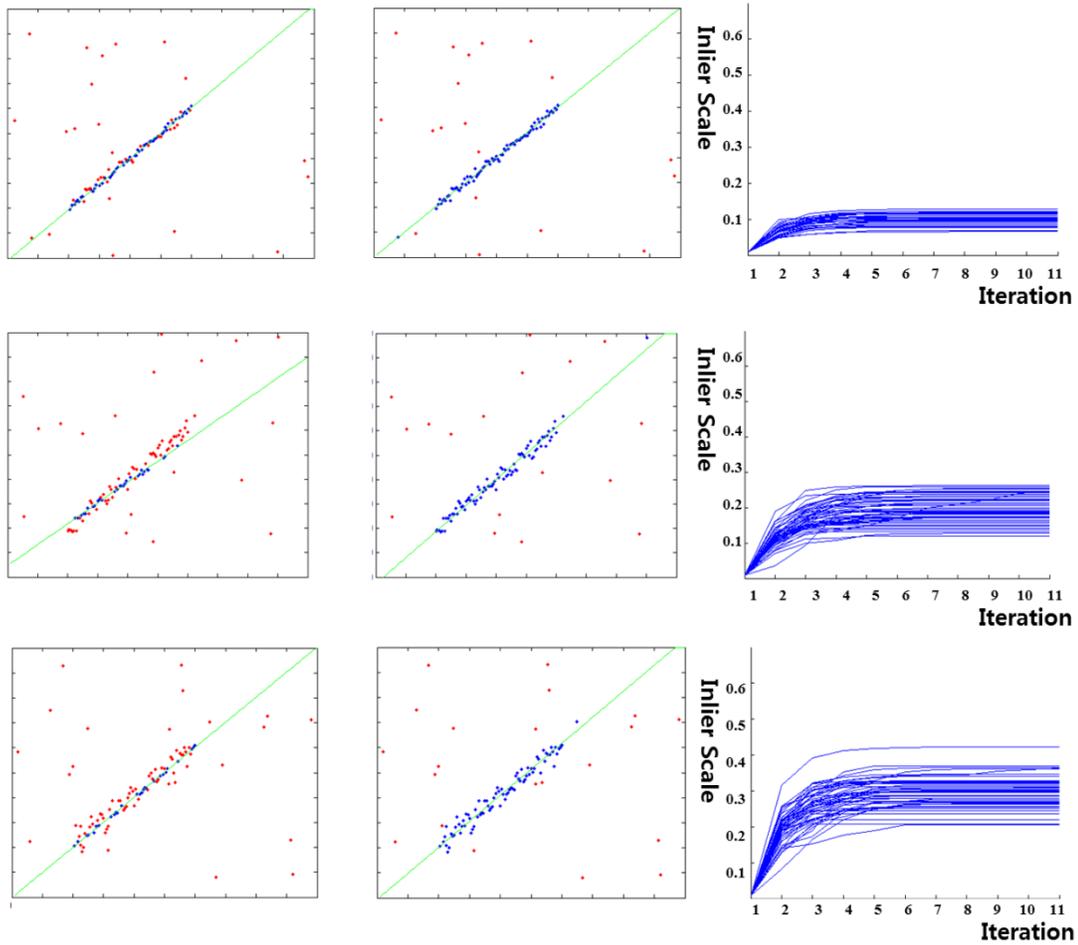

**Fig. 4.** Line fitting results: (first column) example of the results with initial inlier scale, (second column) example of the results with final inlier scale and (third column) transitions of inlier scales during iteration.

We performed the DLT-based IMaxFS-ISE algorithm to fit a 2D line to data with different inlier scales. We set the initial inlier scale $s_0$ to 0.01. The data include 80 inliers with Gaussian noise and 20 gross outliers. For 3 different Gaussian noise levels of 0.1, 0.2 and 0.3, 50 experiments were carried out with random outliers. Figure 4 shows one set of fitting results out of 50 with the initial inlier scale $s_0$ (first column), a set with the finally estimated inlier scale (second column), and estimation of inlier scales over iterations for the 50 experiments (third column). The blue points indicate estimated inliers and red points do outliers. It can be seen that the inlier scales converge well only after several iterations.

### 4.2. IMaxFS-ISE-SU : Homography and Affine Fundamental Matrix Estimation

We performed the DLT-based IMaxFS-ISE-SU to estimate planar homography and affine fundamental matrix for each data subset. For our IMaxFS-ISE-SU algorithm, the Big-M value in Equation 2 was set to 10000, the initial $K^{(0)}$ value and $I_{th}^{In}$ were set to the fixed value of 10.

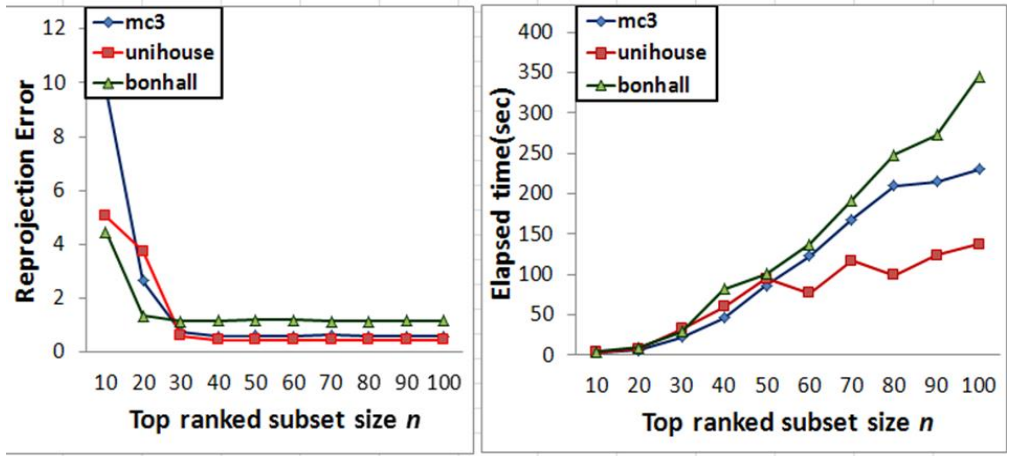

(a)  (b)

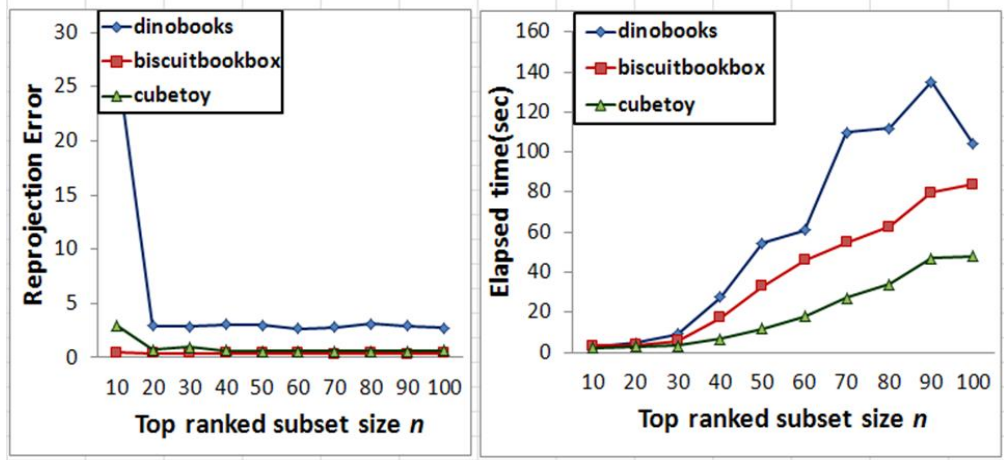

(c)  (d)

**Fig. 5.** Re-projection errors and the elapsed time for computation for subset size n = 10~100

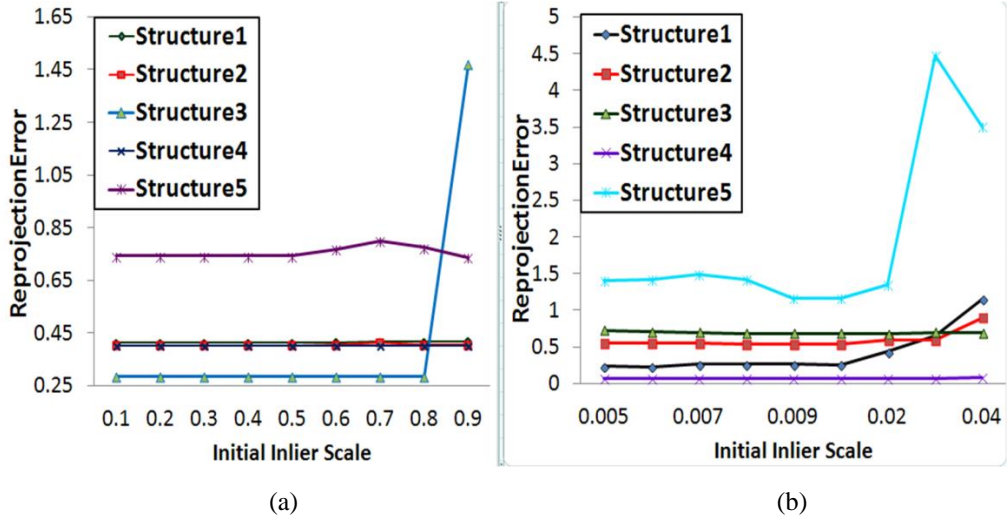

**Fig. 6.** Re-projection errors with varying initial inlier scales from: (a) homography estimation and (b) affine fundamental matrix estimation

To determine the parameters $n$ (subset size) and $s_0$, we experimentally examined the effects of parameter $n$ (subset size) on re-projection error. We investigated the effect of $n$ in the range of [10 100] on data with different outlier rates. Figures 5 (a) and (b) show the re-projection errors and the computation time for homography estimation with the IMaxFS-ISE-SU method, respectively. Only the results from three datasets are shown in the plots. Figures 5 (c) and (d) show the re-projection errors and the computation time for affine fundamental matrix estimation, respectively. As seen in Figure 5, high accuracy is achieved for the subset size $n$ from about 30 up and the computation time gradually increases with $n$. For the IMaxFS-ISE-SU algorithm, we set $n$ to 30 to attain both accuracy and computational efficiency.

We investigated the effect of initial inlier scale $s_0$ on re-projection error for five different structures in three datasets. Figures 6 (a) and (b) show the results for homography and affine fundamental matrix which demonstrate that our algorithm is stable for a wide range of the initial inlier scale $s_0$. If $s_0$ is too small, the computation time becomes too long since our method performs the MaxFS algorithm increasing the inlier scale gradually until $I^{In}_t > I^{In}_{th}$. We find that the $s_0$ values of 0.5 and 0.01 for the estimation of homograpy and fundamental matrix, respectively, are the good compromises between stability and computational efficiency for all the datasets we tested.

To demonstrate the effectiveness of inlier scale estimation we performed the IMaxFS-ISE algorithm on 6 sets of 70 data points, 3 sets for homography and 3 sets for fundamental matrix, and the results are shown in Figures 7 and 8, respectively. The blue squares signify inliers and the yellow crosses do outliers. In each figure, the left images show the inliers estimated from IMaxFS-ISE at the first iteration and the right images show the final results. The estimated inlier scales are at the iteration stages are shown in Figure 9. Note that the structures have different inlier scales since the accuracy of the data points varies depending on the quality of image features extracted from different structures/datasets. We set initial inlier scale $s_0$ to be a small value to avoid breakdown caused by gross outliers or inliers in other structures. If initial inlier scale $s_0$ is too large, the IMaxFS-ISE algorithm may not provide good results since initial hypothesis can be badly biased.

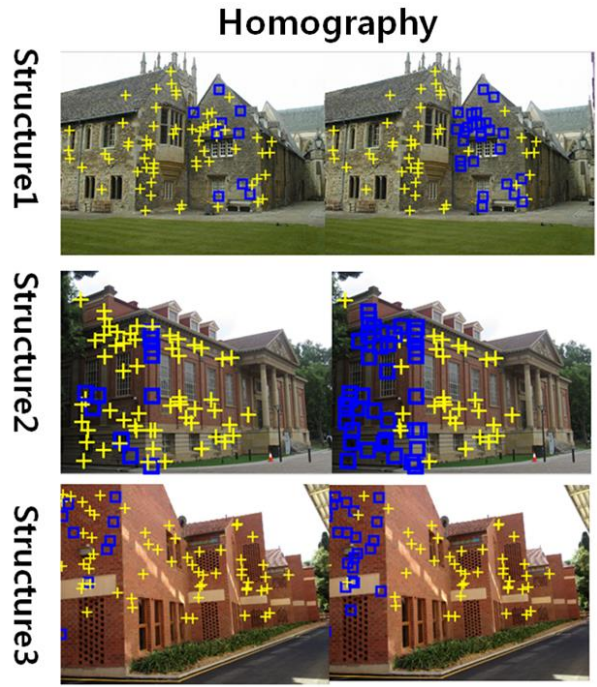

**Fig. 7.** Inlier estimation with IMaxFS-ISE : (left) results with the initial inlier scale $s_0$ and (right) results with the final inlier scale

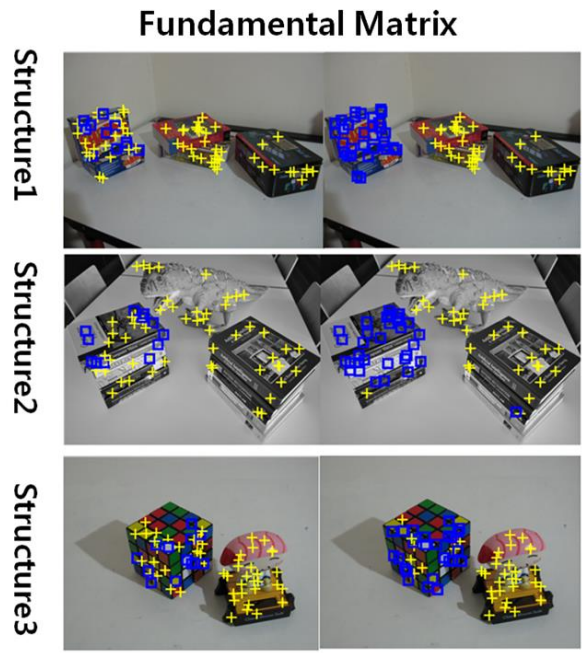

**Fig. 8.** Inlier estimation with IMaxFS-ISE : (left) results with the initial inlier scale $s_0$ and (right) results with the final inlier scale

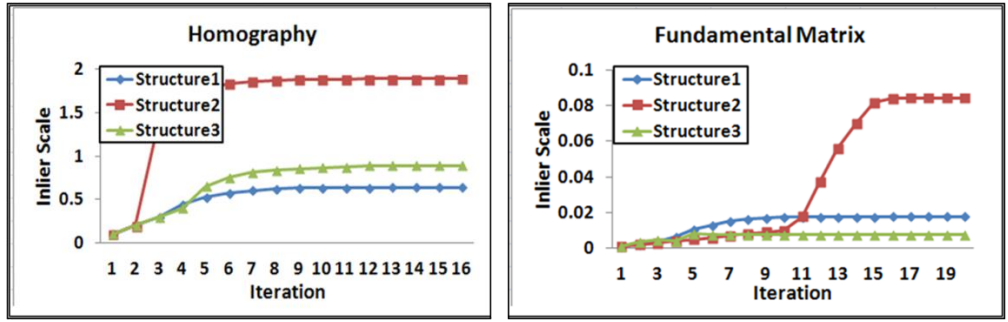

**Fig. 9.** Transition of inlier scales during iteration for each structure in Figure 7 and Figure 8.

### 4.3. Results for Updating Top-$n$ Ranked Subset

We performed our IMaxFS-ISE-SU algorithm on the the *mc3* dataset for homography and on the *biscuitbookbox* dataset for fundamental matrix estimation and the results are shown in Figures 10 and 11, respectively. After finding the first structure by updating the subset of top-30 ranked data points the algorithm removes the data points that belongs to the structure and works on the next top-30 subset for the second structure, and so on. Each row Figures 10 and 11 shows all the iteration stages for a structure. In most cases, subset updating stops under three iterations. The cyan squares indicate inliers in the subset and yellow squares denote gross outliers in the subset. As the quality of the hypothesis is refined in each iteration stage, the inliers for each structure are increased progressively in the subset. Figure 12 shows final fitting results for datasets shown in Figure 10 and Figure 11.

The correspondences from true outliers in a whole dataset mostly rank low and only a few can be present the initial top-$n$ ranked subset. On the other hand, strong matches from pseudo-outliers are naturally included. When the number of structures is large, the inliers from one structure may be insufficient in the initial subset and the initial hypothesis can severely deviate from the true structure. When this happens, our algorithm updates the subset using Equation 8 until it includes enough inliers from a structure

### 4.4. Comparison with Random Sampling Approaches

Our algorithm is compared with four other methods based on random sampling: uniform random sampling (RANSAC) [32, 3], PROSAC [9], Multi-GS [11, 33] and the state-of-the-art algorithm RCM [15, 37]. We implemented the PROSAC algorithm in MATLAB. For performance evaluation, we measured elapsed computation time and the number of generated hypotheses (L) and computed the re-projection errors (mean and standard deviation). The results for the five algorithms are summarized in Tables 1 and 2 with the best results boldfaced.

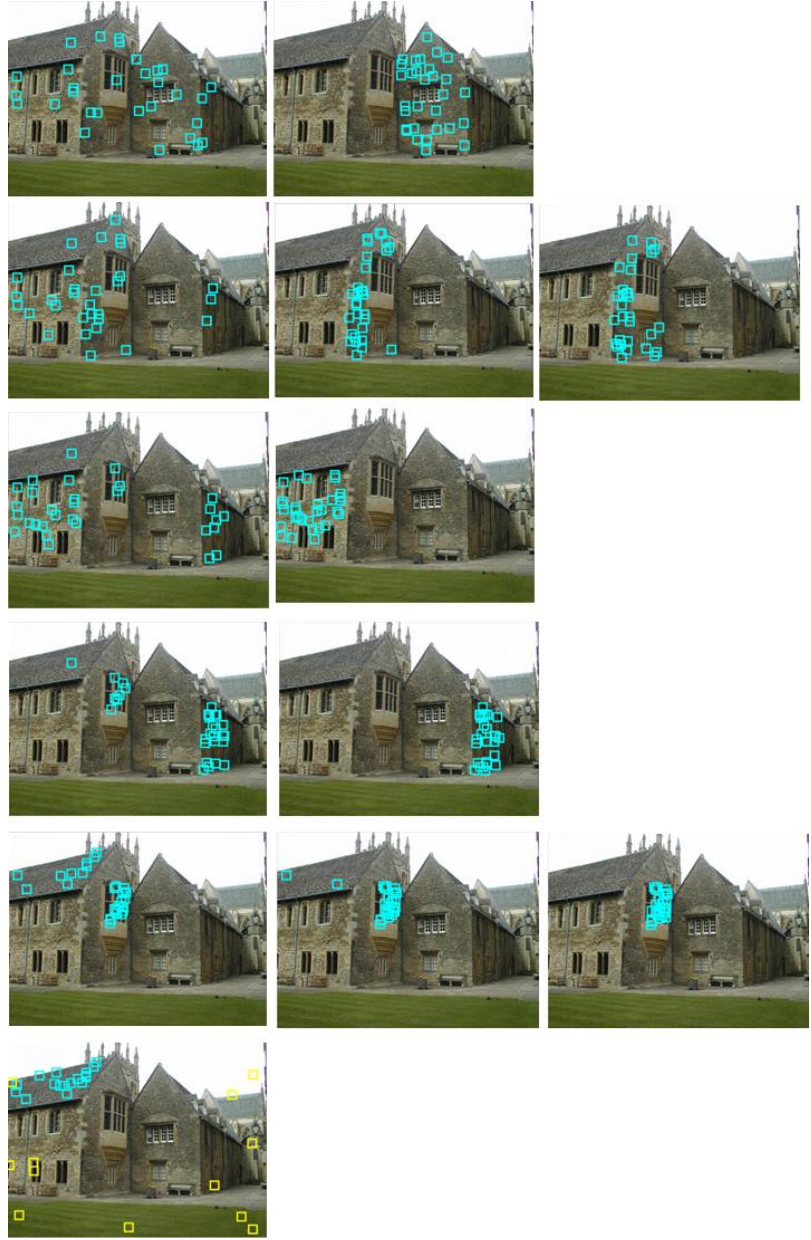

**Fig. 10.** Top-30 subset from the *mc3* dataset: (each row) Updated subset for each structure during iteration.

For each label/structure, the hypothesis that shows the minimum re-projection error is selected. Overall error was calculated by averaging the re-projection errors for all the structures. For RANSAC, PROSAC and Multi-GS, 50 random sampling runs were carried out. The (elapsed) computation times for these three algorithms and ours were made similar. Since our method and RCM run till completion of algorithm, the elapsed computation times for our method and RCM were not limited but measured. For RCM, the average of computation times are measured.

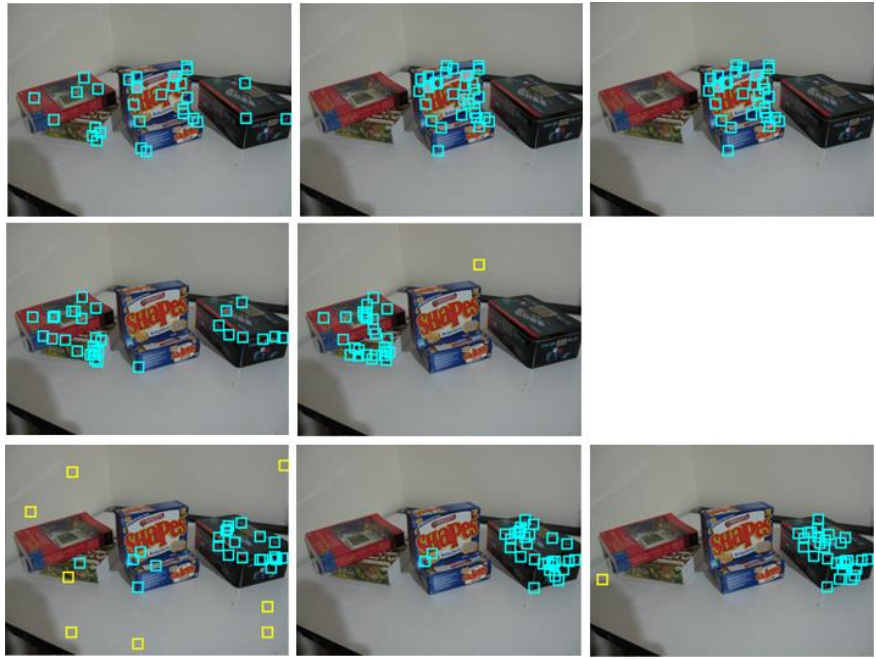

**Fig. 11.** Top-30 subset from the *biscuitbookbox* dataset: (each row) Updated subset for each structure during iteration.

**Homography Estimation**   We tested the performance of our method for estimating planar homographies on real image data. Table 1 summarizes the performance of various methods for each dataset. The outcomes show that our method yields reliable and consistent results with reasonable computational efficiency. In most cases, our method offers the same or higher performance from considerably fewer hypotheses than the RANSAC and the PROSAC algorithms. Since our algorithm generates slightly more hypotheses than the number of structures, there is no need to reduce or merge the hypotheses generated. The RCM yielded largest maximum error and standard deviation.

**Affine Fundamental Matrix Estimation**   We also tested the performance of our method for estimating affine fundamental matrix on real image data. Table 2 shows the performance of various methods for each dataset. Like homography estimation, the results clearly show that our algorithm effectively generates high-quality hypotheses. Note that the random sampling-based methods produces substantial variation in their results.

Figure 13 shows the re-projection errors produced by the five methods on the BCD data as outlier ratio is increased. Our algorithm outperforms the other algorithms as outlier ratio is greatly increased. As the outlier ratio increases, the probability of hitting an all-inlier subset decreases with the random sampling-based approaches and the re-projection errors increases. On the other hand, the hypothesis generated by the IMaxFS-ISE algorithm is little influenced by the outlier ratio.

Figures 14 and 15 show the number of structures that the random sampling methods and ours detect for the estimation of homography and fundamental matrix, respectively. For the raglan dataset with 11 true structures, 10% outliers and the running time of 70 seconds, RANSAC, PROSAC and Multi-GS detect all the 11 structures for homography only in 36, 29 and 2 runs out of 50, respectively. For the dinobooks dataset with 3 true structures, 70% outliers and the running time of 20 seconds, RANSAC, PROSAC and Multi-GS detect all the 3 structures for fundamental matrix only in 13, 28 and 9 runs out of 50, respectively. With a single run, it is hard to trust the results. For the same running times, our method finds all the true structures from all the datasets.

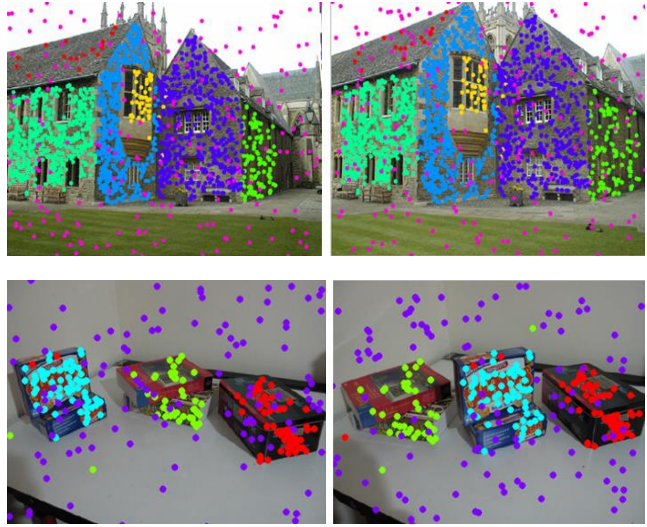

**Fig. 12**. Final fitting results: (top row) the *mc3* dataset and (bottom row) the *biscuitbookbox* dataset.

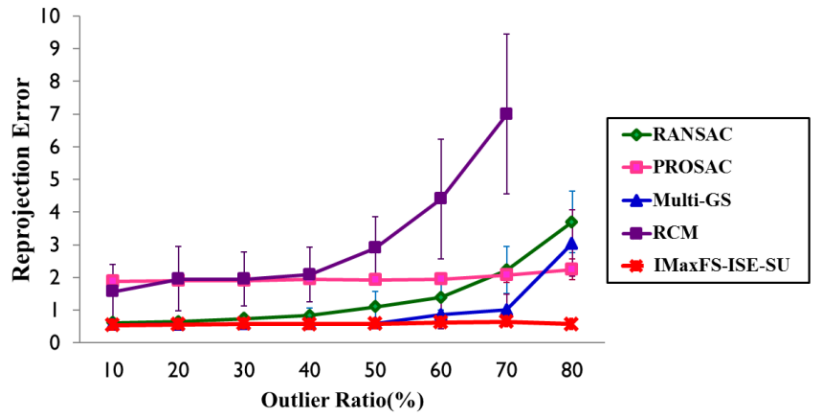

**Fig. 13.** Re-projection errors with varying outlier rates (BCD data)

## 5. Discussion and Conclusion

Despite the recent progresses in multiple-structure fitting, the random sampling-based approaches have the limitation that the number of iterations may not be determined easily without prior information about the data. If the number of iterations is insufficient or outlier ratio is high, the results are inconsistent due to the randomized nature. We present a novel deterministic method partially based on a global optimization technique for multiple-structure model fitting. The MaxFS method can estimate model parameters regardless of outlier ratios as long as the minimum number of inliers are provided. For the method to be effective and computationally tractable, however, inlier scales should be provided and the data size has to be reasonably small. We provide a way of estimating inlier scales and solve the MaxFS problems with only subsets of top-$n$ ranked data points in terms of matching quality.

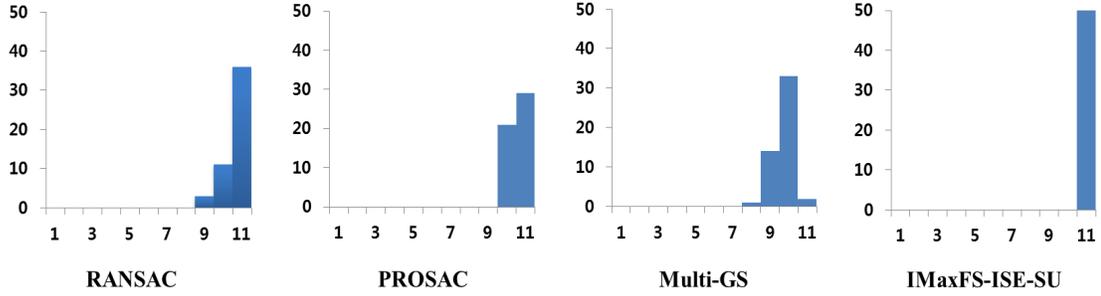

**Fig. 14.** Histograms for the number of structures detected in the estimation of homography. For the raglan dataset with 11 true structures, 10% outliers and the running time of 70 seconds, RANSAC, PROSAC and Multi-GS detect all the 11 structures only in 36, 29 and 2 runs out of 50, respectively. Our method detects all the structures in a single run.

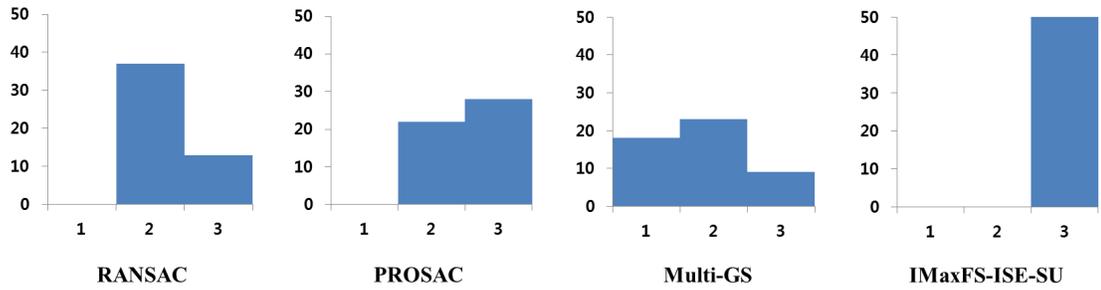

**Fig. 15.** Histograms for the number of structures detected in the estimation of fundamental matrix. For the dinobooks dataset with 3 true structures, 70% outliers and the running time of 20 seconds, RANSAC, PROSAC and Multi-GS detect all the 3 structures only in 13, 28 and 9 runs out of 50, respectively. Our method detects all the structures in a single run.

The presented algorithm, called IMaxFS-ISE-SU, is developed for the estimation of homography and fundamental matrix models. It iteratively estimates the model parameters and inlier scale, generates hypothesis with the top-$n$ ranked subsets selected based on inlier probabilities. The model is refined by updating subset iteratively based on the inlier probabilities which are calculated using matching scores and data fitting residuals. Even in the case that the hypothesis generated from initial subset is badly biased, the algorithm improves the quality of subset using the inlier probabilities. All of the true structures are extracted eventually after a few sequential "fitting-and-removing" procedures. Experimental results show that the IMaxFS-ISE-SU algorithm can generate more reliable and consistent hypotheses than the random sampling-based methods for estimating multiple structures especially when outlier ratio is high. It works robustly without prior knowledge such as inlier ratio, inlier scale and the number of structures.

Our future work includes investigating the use of spatial coherence in addition to matching scores to compute the inlier probabilities. Matching scores are useful for excluding gross outliers but not the strongly matching pseudo-outliers from multiple structures. It may be interesting to see how much helpful spatial coherence is for ruling out some of those pseudo-outliers.

**Table 1.** Performance of various methods on homography estimation for several real datasets.

| Data | OR (%) | Method | RANSAC | PROSAC | Multi-GS | RCM | IMaxFS-ISE-SU |
|---|---|---|---|---|---|---|---|
| raglan | 10% | Elapsed time [sec] | 70 | 70 | 70 | **8.352** | 69.5471 |
| | | Max Error | 0.9328 | 1.4633 | 1.1265 | 1.5783 | **0.8178** |
| | | Std | 0.0761 | 0.0649 | 0.0942 | 0.7225 | **0** |
| | | L | 30014 | **30577** | 1410 | 9 | 11 |
| mc3 | 10% | Elapsed time [sec] | 25 | 25 | 25 | **4.6127** | 24.3706 |
| | | Max Error | 1.2577 | 1.3232 | 1.7618 | 2.239 | **0.4598** |
| | | Std | 0.0927 | 0.079 | 0.232 | 0.2099 | **0** |
| | | L | **15052** | 12815 | 985 | 5 | 7 |
| library | 70% | Elapsed time [sec] | 20 | 20 | 20 | **1.6123** | 19.15 |
| | | Max Error | 1.3715 | 0.9612 | 1.1453 | 6.3131 | **0.8658** |
| | | Std | 0.2193 | 0.0145 | 0.1706 | 3.4552 | **0** |
| | | L | **17749** | 17604 | 2031 | 5 | 3 |
| unihouse | 30% | Elapsed time [sec] | 20 | 20 | 20 | **7.4608** | 20.42 |
| | | Max Error | 1.8277 | 1.4581 | 2.131 | 7.102 | **0.4621** |
| | | Std | 0.2855 | 0.0828 | 0.2641 | 2.6775 | **0** |
| | | L | 8748 | **10520** | 789 | 4 | 6 |
| bonhall | 30% | Elapsed time [sec] | 35 | 35 | 35 | **6.1367** | 36.22 |
| | | Max Error | 1.3054 | **0.7612** | 1.8405 | 4.5943 | 1.274 |
| | | Std | 0.2076 | 0.0273 | 0.3292 | 2.7598 | **0** |
| | | L | **21056** | 20390 | 1327 | 6 | 6 |

**Table 2.** Performance of various methods on affine fundamental matrix estimation for several real datasets.

| Data | OR (%) | Method | RANSAC | PROSAC | Multi-GS | RCM | IMaxFS-ISE-SU |
|---|---|---|---|---|---|---|---|
| cubetoy | 20% | Elapsed time [sec] | 10 | 10 | 10 | **1.0403** | 7.244 |
| | | Max Error | 0.6258 | 0.7684 | 0.6205 | 0.9741 | **0.6127** |
| | | Std | 0.0043 | 0.0493 | 0.0044 | 0.1155 | **0** |
| | | L | 13263 | **13586** | 2067 | 3 | 2 |
| carsbus | 40% | Elapsed time [sec] | 10 | 10 | 10 | **3.8849** | 9.5698 |
| | | Max Error | **0.6793** | 0.7101 | 0.7002 | 1.9509 | 0.7678 |
| | | Std | 0.0016 | 0.0037 | 0.0191 | 0.453 | **0** |
| | | L | **6410** | 6238 | 1171 | 3 | 3 |
| dinobooks | 70% | Elapsed time [sec] | 20 | 20 | 20 | **11.6536** | 17.187 |
| | | Max Error | 3.4369 | 3.8895 | 3.1378 | 6.7627 | **2.7146** |
| | | Std | 0.3924 | 0.3497 | 0.4198 | 1.4298 | **0** |
| | | L | **9899** | 9163 | 1453 | 2 | 3 |
| 4B | 70% | Elapsed time [sec] | **15** | **15** | **15** | 64.5312 | 16.941 |
| | | Max Error | 1.519 | 1.3524 | 1.2634 | 5.6797 | **1.0818** |
| | | Std | 0.1606 | 0.0678 | 0.1605 | 2.1612 | **0** |
| | | L | **2504** | 2500 | 668 | 1 | 6 |
| 5B | 10% | Elapsed time [sec] | 10 | 10 | 10 | **4.4897** | 12.152 |
| | | Max Error | 0.2991 | 1.0414 | **0.2673** | 2.1372 | 0.3195 |
| | | Std | 0.0112 | 0.1273 | 0.0059 | 0.6388 | **0** |
| | | L | **1639** | 1442 | 492 | 3 | 6 |